\pdfoutput=1
\documentclass[11pt,a4paper]{article}
\usepackage{authblk}
\usepackage[hyperref]{emnlp-ijcnlp-2019}
\usepackage{times}
\usepackage{latexsym}
\usepackage[fleqn]{amsmath}
\usepackage{graphicx}
\usepackage{epstopdf}
\usepackage{xcolor}
\usepackage[fleqn]{amsmath}
\usepackage{multirow,multicol}
\usepackage{xytree}
\usepackage{caption}

\definecolor{mygreen}{RGB}{35, 115, 196}
\definecolor{myyellow}{RGB}{186, 115, 0}

\newenvironment{astfootnotes}
  {\par\edef\savedfootnotenumber{\number\value{footnote}}
   
   \setcounter{footnote}{0}}
  {\par\setcounter{footnote}{\savedfootnotenumber}}

\usepackage{url}

\aclfinalcopy 

\title{BookQA: Stories of Challenges and Opportunities}

\author[1*]{\textbf{Stefanos Angelidis}}
\author[2]{\textbf{Lea Frermann}}
\author[3]{\textbf{Diego Marcheggiani}}
\author[3]{\textbf{Roi Blanco}}
\author[3]{\textbf{Llu\'{i}s M\`{a}rquez}}
\affil[1]{Institute for Language, Cognition and Computation, School of Informatics, University of Edinburgh}
\affil[2]{School of Computing and Information Systems, The University of Melbourne}
\affil[3]{Amazon Research}
\affil[ ]{\texttt{s.angelidis@ed.ac.uk \quad\quad lea.frermann@unimelb.edu.au}}
\affil[ ]{\texttt{\{marchegg,roiblan,lluismv\}@amazon.com}}

\date{}

\begin{document}

\maketitle

\begin{abstract}
We present a system for answering questions based on the full text of books (BookQA), which first selects book passages given a question at hand, and then uses a memory network to reason and predict an answer. To improve generalization, we pretrain our memory network using artificial questions generated from book sentences. We experiment with the recently published NarrativeQA corpus, on the subset of \textsl{Who} questions, which expect book characters as answers. We experimentally show that BERT-based retrieval and pretraining improve over baseline results significantly. At the same time, we confirm that NarrativeQA is a highly challenging data set, and that there is need for novel research in order to achieve high-precision BookQA results. We analyze some of the bottlenecks of the current approach, and we argue that more research is needed on text representation, retrieval of relevant passages, and reasoning, including commonsense knowledge.
\end{abstract}

\section{Introduction}
\label{sec:intro}

\begin{astfootnotes}
\footnotetext[1]{Work done while first author was interning at Amazon.}
\end{astfootnotes}

Considerable volume of research work has looked into various Question Answering 
(QA) settings, ranging from retrieval-based QA \cite{voorhees2001trec} to recent
neural approaches that reason over Knowledge Bases (KB)
\cite{bordes2014question}, or raw text
\cite{shen2017reasonet,deng2018read,min2018efficient}. In this paper we use the NarrativeQA corpus \cite{kocisky2018narrativeqa} as a starting point and focus on the task of answering questions from the full text of books, which we call BookQA. BookQA has unique characteristics which prohibit the direct application of current QA methods. For instance, (a) books are usually orders of magnitude longer than the short texts (e.g., Wikipedia articles) used in neural QA architectures; (b) many facts about a book story are never made explicit, and require external or commonsense knowledge to infer them; (c) the QA system cannot rely on pre-existing KBs; (d) traditional retrieval techniques are less effective in selecting relevant passages from self-contained book stories \cite{kocisky2018narrativeqa}; (e)~collecting human-annotated BookQA data is a significant challenge; (f) stylistic disparities in the language used among different books may hinder generalization.

Additionally, the style of book questions may vary significantly, with different approaches being potentially useful for different question types: from queries about story facts that have entities as answers (e.g., \textsl{Who} and \textsl{Where} questions); to open-ended questions that require the extraction or generation of longer answers (e.g., \textsl{Why} or \textsl{How} questions). The difference in reasoning required for different question types can make it very hard to draw meaningful conclusions. 

For this reason, we concentrate on the task of answering~\textsl{Who} questions, which expect book characters as answers (e.g., \textit{``Who is Harry Potter's best friend?''}). This task allows to simplify the output and evaluation (we look for entities, and we can apply precision-based and ranking evaluation metrics), but still retains the important elements of the original NarrativeQA task, i.e., the need to explore over the full content of the book and to reason over a deep understanding of the narrative. Table \ref{tab:examples} exemplifies the diversity and complexity of \textsl{Who} questions in
the data, by listing a set of questions from a single book, which require increasingly complex types of reasoning.

\begin{table}
    \centering
    \small
    \begin{tabular*}{\columnwidth}{l}
    \hline
    \textit{Who is Emily in love with?} \\
    \textit{Who is Emily imprisoned by?} \\
    \textit{Who helps Emily escape from the castle?} \\
    \textit{Who owns the castle in which Emily is imprisoned?} \\
    \textit{Who became Emily's guardian after her father's death?} \\
    \hline
    \end{tabular*}
    \caption{\textsl{Who} questions from NarrativeQA for the book \textit{The Mysteries of Udolpho}, by Ann Radcliffe. The diversity and complexity of questions in the corpus remains high, even when considering only the subset of \textsl{Who} questions that expect characters as answers.}
    \label{tab:examples}
\end{table}
 
NarrativeQA~\cite{kocisky2018narrativeqa} is the first publicly available dataset for QA over long narratives, namely the full text of books and movie scripts. The full-text task has only been addressed by \citet{tay-etal-2019-simple}, who proposed a curriculum learning-based two-phase approach (\textsl{context selection} and \textsl{neural inference}). More papers have looked into answering NarrativeQA's questions from only book/movie \textsl{summaries} \cite{indurthi-etal-2018-cut,bauer-etal-2018-commonsense,tay-etal-2018-multi,tay-et-al-neurips2019,nishida-etal-2019-multi}. 
This is a fundamentally simpler task, because: i) the systems need to reason over a much shorter context, i.e., the summary; and ii) there is the certainty that the answer can be found in the summary. 
This paper is another step in the exploration of the full NarrativeQA task, and embraces the goal of finding an answer in the complete book text. We propose a system that first selects a small subset of relevant book passages, and then uses a memory network to reason and extract the answer from them. The network is specifically adapted for generalization across books. We analyze different options for selecting relevant contexts, and for pretraining the memory network with artificially created question--answer pairs. Our key contributions are: i) this is the first systematic exploration of the challenges in full-text BookQA, ii) we present a full pipeline framework for the task, iii) we publish a dataset of \textsl{Who} questions which expect book characters as an answer, and iv) we include a critical discussion on the shortcomings of the current QA approach, and we discuss potential avenues for future research.

\section{Book Character Questions}
\label{sec:corpus}

NarrativeQA was created using a large annotation effort, where participants
were shown a human-curated \textsl{summary} of a book/script and were asked to
produce question-answer pairs \textsl{without referring to the full story}. The
main task of interest is to answer the questions by looking at the \textsl{full
story} and not at the summary, thus ensuring that answers cannot be simply copied
from the story. The full corpus contains 1,567 stories (split equally between books
and movies) and 46,765 questions.

We restrict our study to \textsl{Who} questions
about \textsl{books}, which have \textsl{book characters} as answers (e.g.,
\textit{``Who is charged with attempted murder?''}). Using the book preprocessing 
system, book-nlp (see Section \ref{sec:booknlp}), and a combination of automatic and crowdsourced efforts, we obtained a total of 3,427 QA pairs,
spanning 614 books.\footnote{To obtain the BookQA data, follow the instructions at: \url{https://github.com/stangelid/bookqa-who}.}


\begin{figure*}[t]
  \centering
  \begin{minipage}{\textwidth}
    \includegraphics[width=\textwidth]{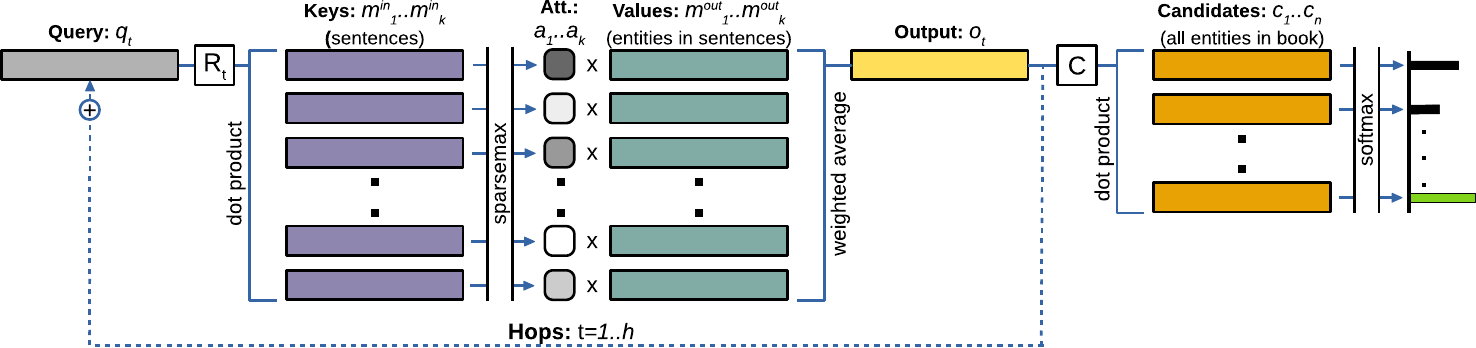}
  \end{minipage}\vspace{1mm}
  \fbox{
  \small
  \begin{minipage}{.37\textwidth}
    \textbf{Initialization:}\vspace{-2mm}
    \begin{align*}
      \mathsf{Query\!:}&\;\mathrm{q}_{t=0} = \mathit{avg}(\mathrm{v}_{qw_1}, \dots, \mathrm{v}_{qw_m}) \\
      \mathsf{Keys\!:}&\;\mathrm{m^{in}_i} = \mathit{avg}(\mathrm{v}_{sw_1}, \dots, \mathrm{v}_{sw_l}) \\
      \mathsf{Values\!:}&\;\mathrm{m^{out}_i} = \mathit{avg}(\mathrm{v}_{c_1 \in s}, \dots) \\
      \mathsf{Candidates\!:}&\;\mathrm{c_j} = \mathrm{v}_{c_j}
    \end{align*}
  \end{minipage}%
  \begin{minipage}{.30\textwidth}
    \textbf{At Hop t:}\vspace{-2mm}
    \begin{align*}
      &a_{ti} = \mathit{sparsemax}(\mathrm{q}_t\mathrm{R}_t\mathrm{m_i}) \\
      &o_t = \sum_i a_{ti}\mathrm{m^{out}_i} \\ 
      &q_{t+1} = q_t + o_t
    \end{align*}
  \end{minipage}%
  \begin{minipage}{.30\textwidth}
    \textbf{After last hop:}\vspace{+2mm}
    \begin{align*}
      &p(c_j) = softmax(o_hC\mathrm{v}_{c_j})
    \end{align*}\vspace{+3mm}
  \end{minipage}
  }
  \caption{Overview of our Key-Value Memory Network for BookQA. Encodings of
    questions, keys (selected sentences), and values (characters mentioned in those
    sentences) are loaded. After multiple hops of inference, the model's output is
    compared against the candidate answers' encodings to make a prediction.}
  \label{fig:memnet}
\end{figure*}

\section{BookQA Framework}

The length of books and limited annotated data prohibit the application of end-to-end neural
QA models that reason over the full text of a book. Instead, we opted for a pipeline approach, 
whose components are described below.

\subsection{Book \& Question Preprocessing} 
\label{sec:booknlp}

Books and questions are preprocessed in advance using the book-nlp parser \cite{bamman2014bayesian},
a system for character detection and shallow parsing in books
\cite{iyyer2016feuding,frermann2017inducing} which provides, among others:
sentence segmentation, POS tagging, dependency parsing, named entity recognition, 
and coreference resolution.
The parser identifies and clusters character mentions, so that all coreferent (direct or pronominal) character mentions are associated with the same unique character identifier.

\subsection{Context Selection}
In order to make inference over book text tractable and give our model a better chance at predicting the correct answer, we must restrict the context to only a small number of book sentences. We developed two context selection methods to retrieve relevant book
passages, which we define as windows of 5 consecutive sentences:\\  \vspace{-3mm}

\noindent \textbf{IR-style selection (BM25F):} We constructed a searchable \textsl{book
    index} to store individual book sentences. We replace every book character mention, 
    including pronoun references, with the character's unique identifier. 
    At retrieval time, we similarly replace character mentions in each question, and rank passages from the corresponding book using BM25F 
    \cite{zaragoza2004microsoft}.\\ \vspace{-3mm}

\noindent \textbf{BERT-based selection:} We developed a neural context selection method,
    based on the BERT language representation model \cite{devlin2018bert}. 
    A pretrained BERT model is fine-tuned to predict if a sentence is relevant to a 
    question, using positive (\textsl{questions, summary sentence}) training pairs 
    which have been heuristically matched. Randomly sampled negative pairs were also used. At
    retrieval time, a question is used to retrieve relevant passages
    from the full text of a book.

\subsection{Neural Inference}
Having replaced character mentions in questions and books with character identifiers, 
we first pretrain word2vec embeddings \cite{mikolov2013distributed} for all words and book characters 
in our corpus.\footnote{Character identifiers are treated like all other tokens.} Our neural inference model is a variant of the Key-Value Memory Network
(KV-MemNet) \cite{miller2016key}, which has been previously applied to QA tasks
over KBs and short texts. The original model was 
designed to handle a fixed set of potential answers across all QA examples, as do most 
neural QA architectures. This comes in contrast with our task,
where the pool of candidate characters is different for each book. Our KV-MemNet variant,
illustrated in Figure~\ref{fig:memnet}, uses a dynamic output layer where different 
candidate answers are made available for different books, while the remaining model
parameters are shared.

A question is initially represented as $\mathrm{q}_0$, i.e., the average of its word
embeddings\footnote{Experiments with more sophisticated question/sentence representation
variants showed no significant improvements.} (gray vector). The \textsl{Key} memories $\mathrm{m^{in}_1} \dots
\mathrm{m^{in}_k}$ (purple vectors) are filled with the $k$ most relevant
sentences, as retrieved from the context selection step, using the average
of their word embeddings. \textsl{Value} memories $\mathrm{m^{out}_1} \dots
\mathrm{m^{out}_k}$ (green vectors) contain the average embedding of all
characters mentioned in the respective sentence, or a padding vector if
no character is mentioned. Candidate embeddings $\mathrm{c}_1 \dots
\mathrm{c}_n$ (orange vectors) hold the embeddings of every character in the
current book. The model makes multiple reasoning hops \mbox{$t=1 \dots h$} over the
memories. At each hop, $\mathrm{q}_t$ is passed through linear layer $R_t$ and
is then compared against all key memories. The \textsl{sparsemax}-normalized
\cite{martins2016softmax} attention weights $a_1 \dots a_k$ are then used for
obtaining output vector $\mathrm{o}_t$, as the weighted average of value
memories. The process is repeated $h$ times, and the
final output is passed through linear layer $C$, before being compared against
all candidate vectors via dot-product, to obtain the final prediction. The model
is trained using negative log-likelihood. \vspace{-1mm}

\begin{figure}[t]
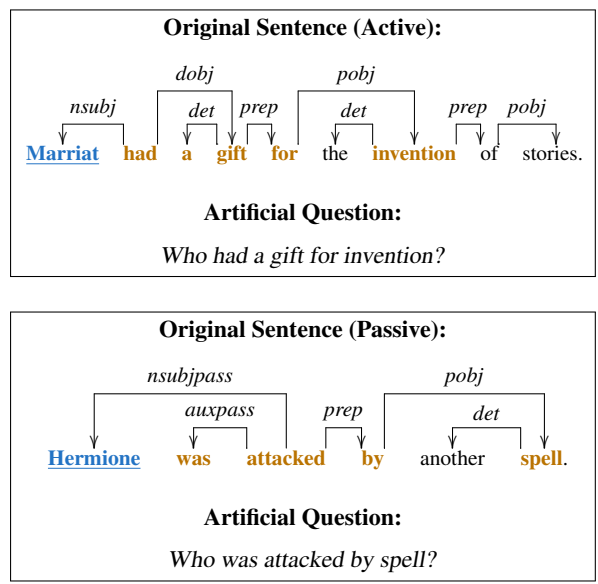

    \small
    \centering
    \fbox{%
    \parbox{.97\columnwidth}{%
    \centering
    \textbf{Original Sentence (Active):}\\\medskip
    \scalebox{0.88}{
    \xytext{
      \xybarnode{\textbf{\textcolor{mygreen}{\underline{Marriat}}}} &~&
      \xybarnode{\textbf{\textcolor{myyellow}{had}}}
        \xybarconnect(UL,U){-2}"_{\footnotesize \textit{nsubj}}"
        \xybarconnect[8](UR,U){4}"^{\footnotesize \textit{dobj}}"
        &~&
      \xybarnode{\textbf{\textcolor{myyellow}{a}}} &~&
      \xybarnode{\textbf{\textcolor{myyellow}{gift}}} 
        \xybarconnect(UL,U){-2}"_{\footnotesize \textit{det}}"
        \xybarconnect(UR,UL){2}"^{\footnotesize \textit{prep}}"
        &~&
      \xybarnode{\textbf{\textcolor{myyellow}{for}}} 
        \xybarconnect[8](UR,U){4}"^{\footnotesize \textit{pobj}}"
        &~&
      \xybarnode{the} &~&
      \xybarnode{\textbf{\textcolor{myyellow}{invention}}} 
        \xybarconnect(UL,U){-2}"_{\footnotesize \textit{det}}"
        \xybarconnect(UR,UL){2}"^{\footnotesize \textit{prep}}"
        &~&
      \xybarnode{of} 
        \xybarconnect(UR,U){2}"^{\footnotesize \textit{pobj}}"
        &~&
      \xybarnode{stories. }
    }}\\
    \bigskip
    \textbf{Artificial Question:}\\\medskip \textsl{Who had a gift for invention?}}}\\
    \bigskip
    \fbox{%
    \parbox{.97\columnwidth}{%
    \centering
    \textbf{Original Sentence (Passive):}\\\medskip
    \scalebox{0.9}{
    \xytext{
      \xybarnode{\textbf{\textcolor{mygreen}{\underline{Hermione}}}} &~~~&
      \xybarnode{\textbf{\textcolor{myyellow}{was}}} &~~~&
      \xybarnode{\textbf{\textcolor{myyellow}{attacked}}}
        \xybarconnect[8]{-4}"_{\footnotesize \textit{nsubjpass}}"
        \xybarconnect(UL,U){-2}"_{\footnotesize \textit{auxpass}}"
        \xybarconnect(UR,UL){2}"^{\footnotesize \textit{prep}}"
        &~~~&
      \xybarnode{\textbf{\textcolor{myyellow}{by}}}
        \xybarconnect[8](UR,U){4}"^{\footnotesize \textit{pobj}}"
        &~~~&
      \xybarnode{another} &~~~&
      \xybarnode{\textbf{\textcolor{myyellow}{spell}}.}
        \xybarconnect(UL,U){-2}"_{\footnotesize \textit{det}}"
    }}\\
    \bigskip
    \textbf{Artificial Question:}\\\medskip \textsl{Who was attacked by spell?}}}
    \caption{Examples of artificial questions generated from the dependency trees of an 
    active voice (top) and a passive voice (bottom) sentence. The correct answer 
    (\textsl{verb's subject}) is marked with
    \textcolor{mygreen}{\textbf{\underline{blue}}}, whereas the \textcolor{myyellow}{\textbf{yellow}}
    words are used in the question. The remaining words are discarded by pruning the
    dependency tree.}
    \label{fig:artif}
\end{figure}
 
\begin{table*}[t]
  \centering
  \small
    \begin{tabular}{l|cc|cc|cc}
      \multicolumn{1}{r|}{\textsl{\textcolor{black!60}{Metric $\to$}}}
                             & \multicolumn{2}{c|}{\textbf{P@1}}
                             & \multicolumn{2}{c|}{\textbf{P@5}}
                             & \multicolumn{2}{c}{\textbf{MRR}} \\
      \multicolumn{1}{r|}{\textcolor{black!60}{\textsl{Context selection $\to$}}}
                             & BM25F & BERT
                             & BM25F & BERT
                             & BM25F & BERT \\
      \hline
      \textbf{Baselines:} & & & & & & \\    
      Book frequency & \multicolumn{2}{c|}{15.73}
                     & \multicolumn{2}{c|}{56.29}
                     & \multicolumn{2}{c}{0.337} \\
      Context frequency & 10.53 & 13.80 & 51.42 & 53.02 & 0.276 & 0.305 \\
      \hline
      \textbf{KV-MemNet:} & & & & & & \\
      No pretraining & 15.57$\pm$0.97 & 15.89$\pm$0.95 & 58.18$\pm$1.57 & 58.77$\pm$1.29 & 0.339$\pm$0.006 & 0.343$\pm$0.008 \\
      Pretrain w/ Artif. Qs & 15.92$\pm$0.73 & \textbf{18.73}$\pm$1.07 & 61.25$\pm$0.74 & \textbf{62.81}$\pm$1.07 & 0.351$\pm$0.005 & \textbf{0.376}$\pm$0.006 \\
      \hline
    \end{tabular}
    \caption{Precision scores (P@1, P@5), and Mean
    Reciprocal Rank (MRR) for frequency-based baselines and our system, with and
    without pretraining. We report average and standard deviation over 50 runs.}
    \vspace{-2mm}
    \label{tab:results}
\end{table*}

\subsection{Pretraining}
\label{sec:pretraining}

A significant obstacle towards effective BookQA is the limited amount of data
available for supervised training. A potential
avenue for overcoming this is pretraining the neural inference model on an
auxiliary task, for which we can generate orders of magnitude more training
examples. To this end, we generated 688,228 artificial questions from the book text using a set of simple pruning rules over the dependency trees of book sentences. We used all book sentences where a character mention is the agent or the patient of an active voice verb, or the patient of a passive voice verb. Two examples are illustrated in Figure \ref{fig:artif}: at the top, the active voice sentence \textit{``Marriat had a gift for the invention of stories.''} is transformed into the question \textit{``Who had a gift for invention?''} and, at the bottom, the passive voice sentence \textit{``Hermione was attacked by another spell.''} is transformed into the question \textit{``Who was attacked by a spell?''}. The previous 20 book sentences, including the source sentence, are used as context during pretraining.

\section{Experimental Setup}
\label{sec:setup}

For every
question, 100 sentences (top 20 passages of five sentences) were selected as
contexts using our retrieval methods. We used word and book character embeddings
of 100 dimensions. The number of reasoning hops was set
to 3. When no pretraining was performed, we trained on the real QA examples for
60 epochs, using Adam with initial learning rate of $10^{-3}$, which we reduced
by 10\% every two epochs. Word and character embeddings were fixed during
training. When using pretraining, we trained the memory
network for one epoch on the auxiliary task, including the embeddings. Then, the
model was fine-tuned as described above on the real QA examples where, again,
embeddings were fixed. We use
Precision at the 1st and 5th rank (P@1 and P@5) and Mean Reciprocal Rank (MRR)
as evaluation metrics.
We adopted a 10-fold cross validation approach and performed 5 trials for each cross validation split, for a total of 50 experiments.\\ \vspace{-3mm}

\noindent \textbf{Baselines:} We implemented a random baseline and two frequency-based baselines, where the most frequent character in the entire book (\textsl{Book} frequency) or the selected context (\textsl{Context} frequency) was selected as the answer.

\begin{figure*}
    \begin{minipage}{.2\textwidth}
        \centering
        \includegraphics[width=\linewidth]{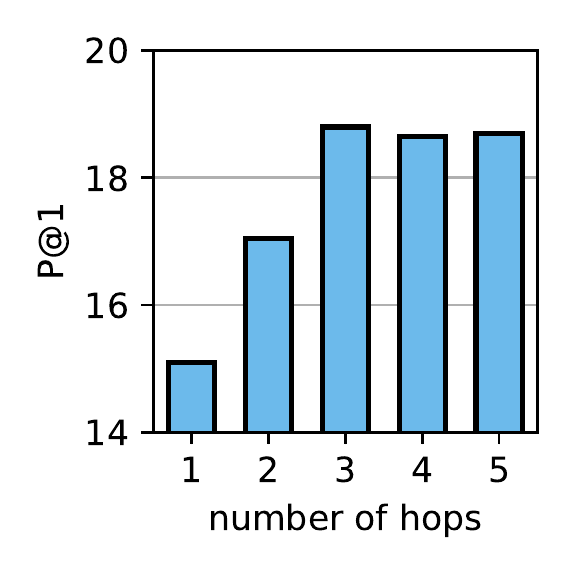}
        \caption{P@1 for different number of hops.}
        \label{fig:hops}
    \end{minipage}%
    \hfill%
    \begin{minipage}{.3\textwidth}
        \centering
        \includegraphics[width=\linewidth]{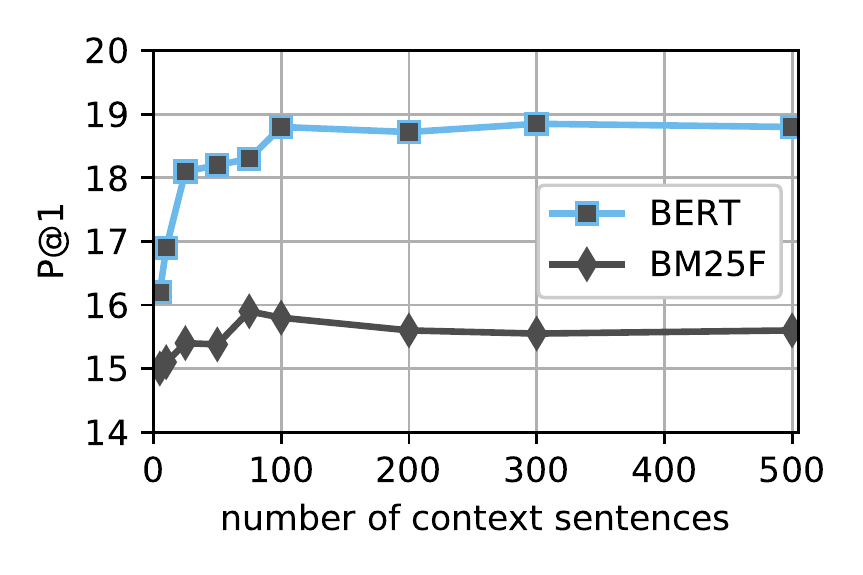}
        \caption{P@1 for varying context sizes from BM25F and BERT.}
        \label{fig:msize}
    \end{minipage}%
    \hfill%
    \begin{minipage}{.46\textwidth}
            \centering
            \small
            \begin{tabular}{p{4.2cm}|l|r}
                \hline
                \multirow{2}{4.2cm}{correct character mentioned in context} & BM25F & 69.7\% \\
                & BERT & 74.7\% \\
                \hline
                full evidence found in context & \multirow{3}{*}{BM25F} & 27\% \\
                partial evidence found in context & & 47\% \\
                no evidence found in context & & 26\% \\
                \hline
            \end{tabular}
            \captionof{table}{Percentage of contexts where the correct character is mentioned (top).
            Percentage of contexts where full/partial/no evidence for the answer was found according 
            to crowd-workers who examined a sample of 100 cases (bottom).}
            \label{tab:evidence}
    \end{minipage}
\end{figure*}

\section{Results} 
\label{sec:results}

Our main results are presented in Table~\ref{tab:results}.
Firstly, we observe one of the dataset's biases, as the book's most frequent
character is the correct answer in more than 15\% of examples, whereas selecting a character at random would only yield the correct answer 2.5\% of the time.

With regards to our BookQA pipeline, the results confirm that BookQA is a very challenging task. Without pretraining, our KV-MemNet which uses IR contexts achieves 15.57\% P@1, and it is slightly outperformed by its BERT-based counterpart.\footnote{Despite the similar performance to the Book frequency baseline, we \textsl{did not} observe that our model was systematically selecting the most frequent character as the answer.} When pretraining the memory network with artificial questions, the BERT-based model achieves 18.73\% P@1. The same trend is observed with the other metrics.\\ \vspace{-3mm}

\noindent \textbf{Number of hops:} We also calculated the impact of the number of hops with respect to the P@1 for a pretrained model fine-tuned with BERT-selected contexts. Figure \ref{fig:hops} shows that performance increases up to 3 hops and then it stabilizes.\\ \vspace{-3mm}

\noindent \textbf{Context size:} We expected the context size (i.e., the number of retrieved sentences that we store in the memory slots of our KV-MemNet) to significantly affect performance. Smaller contexts, obtained by only retrieving 
the topmost relevant passages, might miss important evidence for answering a 
question at hand. Conversely, larger contexts might introduce noise in the form of
irrelevant sentences that hinder inference. Figure \ref{fig:msize} shows the
performance of our method when varying the number of context sentences (or, equivalently, memory slots). The neural inference model struggles for very small context sizes and achieves its best performance for 75 and 100 context sentences obtained by BM25F and BERT, respectively. For both alternatives, we observe no further improvements for larger contexts.\\ \vspace{-3mm}

\begin{figure}[t]
    \centering
    \begin{minipage}{\columnwidth}
        \centering
        \includegraphics[width=\linewidth]{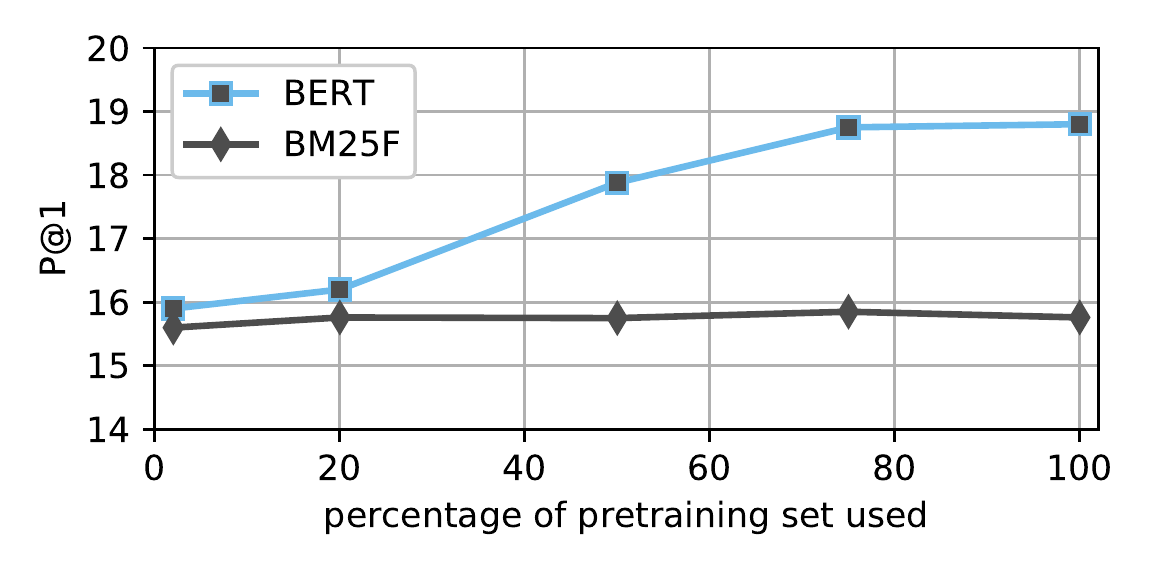}
        \caption{P@1 for varying percentage of pretraining questions used (BM25F and BERT contexts).}
        \label{fig:pre-size}
    \end{minipage}\vspace{3mm}
    \begin{minipage}{\columnwidth}
        \centering
        \includegraphics[width=\linewidth]{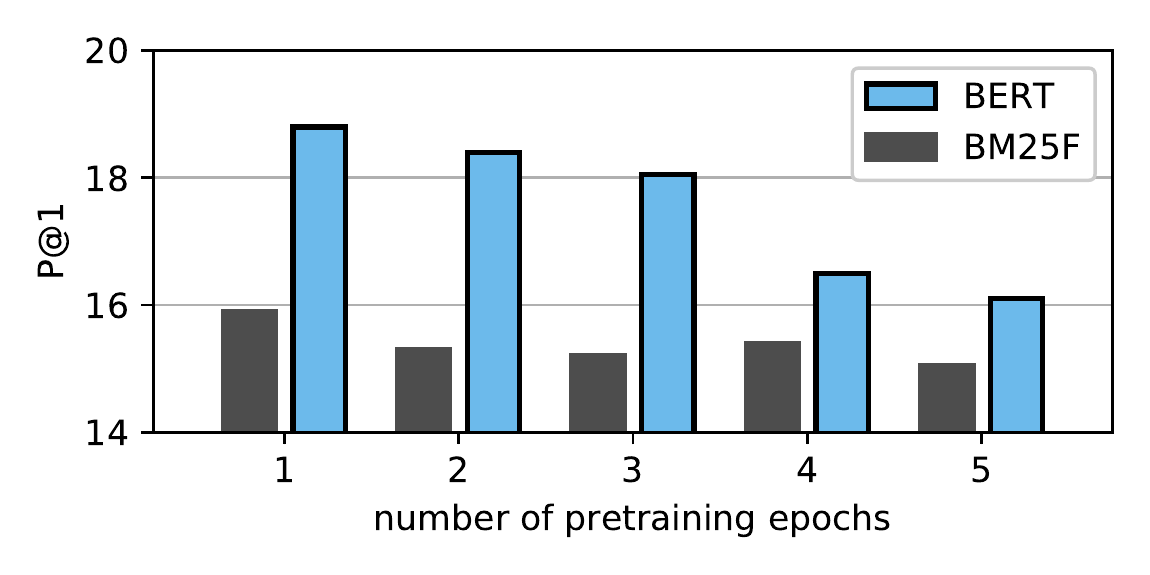}
        \caption{P@1 as a function of pretraining epochs for BM25F and BERT contexts.}
        \label{fig:pre-epochs}
    \end{minipage}
\end{figure}

\noindent \textbf{Pretraining size \& epochs:} A key component of our BookQA 
framework is the pretraining of our neural inference model with artificially 
generated questions. Although it helped achieve the highest percentage of
correctly answered questions, the performance gains were relatively small given the number of artificial questions used to pretrain the model. We further investigated the effect of pretraining by varying the number of 
artificial questions used during training and the number of pretraining epochs. 
Figure \ref{fig:pre-size} shows the QA performance achieved on the real BookQA questions (using BM25F or BERT contexts) after pretraining on a randomly sampled subset of the artificial questions. For our BERT-based  variant, the
pencentage of correctly answered questions increases steadily, but flattens out when reaching 75\% of 
pretraining set usage. On the contrary, when using BM25F contexts we achieved insignificant gains, with performance appearing constrained by the quality of retrieved passages. In Figure \ref{fig:pre-epochs} we show P@1 scores as a function of 
the number of pretraining epochs. Best performance is achieved after only one epoch for both variants, indicating
that further pretraining might cause the model to overfit to the simpler type of reasoning required for answering 
artificial questions.

\subsection{Further Discussion}
\label{sec:discussion}

Despite the limitation to {\sl Who} questions, the employment of strong models for context selection and 
neural inference, and our pretraining efforts, the overall BookQA accuracy remains modest, as our best-performing system achieves a P@1 score below 20\%. 
Even when we only allowed our system to answer if it was very confident (according to the probability difference between top-ranked candidate answers), it answered correctly 35\% of times.

We have identified a number of reasons which inhibit better performance. Firstly, the passage selection process constrains the answers that can be logically inferred. We provide our findings in regards to this claim in Table \ref{tab:evidence}. We calculated that the correct answer appears in the IR-selected contexts in 69.7\% of cases. For BERT-selected contexts it appears in 74.7\% of cases. In practice, however, these upper-bounds are not achievable; even when the correct answer appears in the context, there is no guarantee that enough evidence exists to infer it. To further investigate this, we ran a survey on Amazon Mechanichal Turk, where participants were asked to indicate if the selected context (IR-retrieved) contained partial or full evidence for answering a question. For a set of 100 randomly sampled questions, participants found full evidence for answering a question in just 27\% of cases. Only partial evidence was found in 47\% of cases, and no evidence in the remaining 26\%. 

Manual inspection of context sentences indicated that a common reason for the absence of full evidence
is the inherent vagueness of literary language. Repeated expressions or direct references to character names are often avoided by authors, thus requiring very accurate paraphrase detection and coreference resolution. We 
believe that commonsense knowledge is particularly crucial for improving BookQA. When exploring the output of our system, we repeatedly found cases
where the model failed to arrive at the 
correct answer due to key information being left implicit. Common examples we identified were: i)
character relationships which were clear to the reader, but never explicitly described
(e.g., \textsl{``Who did Mark's best friend marry?"}); ii) the attitude of a character towards an 
event or situation (e.g., \textsl{``Who was angry at the school's policy?"}); iii) the relative 
succession of events (e.g., \textsl{``Who did Marriat talk to after the big fight?"}). The injection
of commonsense knowledge into a QA system is an open problem for general and, consequently, BookQA.

In regards to pretraining, the lack of further improvements is likely related to the difference in the type of reasoning required for answering the artificial questions and the real book questions. By construction, the artificial questions will only require that the model accurately matches the source sentence, without the need for complex or multi-hop reasoning steps. In contrast, real book questions require inference over information spread across many parts of a book. We believe that our proposed auxiliary task mainly helps the model by improving the quality of word and book character representations. It is, however, clear from our results that pretraining is an important avenue for improving BookQA accuracy, as it can increase the number of training instances by many orders of magnitude with 
limited human involvement. Future work should look into automatically constructing auxiliary questions that better approximate the types of reasoning required for realistic questions on the content of books.

We argue that the shortcomings discussed in previous paragraphs, i.e., the lack of evidence in retrieved passages, the difficulty of long-term reasoning, the need for paraphrase detection and commonsense knowledge, and the challenge of useful pretraining, are not specific to \textsl{Who} questions. On the contrary, we expect that the requirement for novel research in these areas will generalize or, potentially, increase in the case of more general questions (e.g., open-ended questions).

\section{Conclusions}
We presented a pipeline BookQA system to answer character-based questions on NarrativeQA, from the full book text. By constraining our study to \textsl{Who} questions, we simplified the task's output space, while largely retaining the reasoning challenges of BookQA, and our ability to draw conclusions that will generalize to other question types. Given a \textsl{Who} question, our system retrieves a set of relevant passages from the book, which are then used by a memory network to infer the answer in multiple hops. A BERT-based trained retrieval system, together with the usage of artificial question-answer pairs to pretrain the memory network, allowed our system to significantly outperform the lexical frequency-based baselines. The use of BERT-retrieved contexts improved upon a simpler IR-based method although, in both cases, only partial evidence was found in the selected contexts for the majority of questions. Increasing the number of retrieved passages did not result in better performance, highlighting the significant challenge of accurate context selection. 
Pretraining on artificially generated questions provided promising improvements, but the automatic construction of realistic questions that require multi-hop reasoning remains an open problem.
These results confirm the difficulty of the BookQA challenge, and indicate that there is need for novel research in order to achieve high-quality BookQA. Future work on the task must focus on several aspects of the problem, including: (a) improving context selection, by combining IR and neural methods to remove noise in the selected passages, or by jointly optimizing for context selection and answer extraction \cite{das2018multistep}; (b) using better methods for encoding questions, sentences, and candidate answers, as embedding averaging results in information loss; (c) pretraining tactics that better mimic the real BookQA task; (d) incorporation of commonsense knowledge and structure, which was not addressed in this paper.

\paragraph{Acknowledgments} We would like to thank Hugo Zaragoza and Alex Klementiev for their valuable insights, feedback and support on the work presented in this paper.

\bibliography{emnlp-ijcnlp-2019}
\bibliographystyle{acl_natbib}

\end{document}